\documentclass{article}

\usepackage{amsmath}
\usepackage{array}
\usepackage{multirow}
\usepackage{caption}

\usepackage{PRIMEarxiv}

\usepackage[utf8]{inputenc} 
\usepackage[T1]{fontenc}    
\usepackage{hyperref}       
\usepackage{url}            
\usepackage{booktabs}       
\usepackage{amsfonts}       
\usepackage{nicefrac}       
\usepackage{microtype}      
\usepackage{lipsum}
\usepackage{fancyhdr}       
\usepackage{graphicx}       
\graphicspath{{media/}}     

\title{Stock Prediction via a Dual Relation Fusion Network incorporating Static and Dynamic Relations
}

\usepackage{orcidlink}
\usepackage{footmisc}

\newcommand{\Author}[2]{
  \textbf{#1}\textsuperscript{#2}%
}

\author{
  \Author{Long Chen}{1} \and
  \Author{Huixin Bai\textsuperscript{*}}{1} \and
  \Author{Mingxin Wang}{1} \and
  \Author{Xiaohua Huang}{2} \and
  \Author{Ying Liu}{2} \and
  \Author{Jie Zhao}{3} \and
  \Author{Ziyu Guan}{4} 
  \\[1em]
  \begin{minipage}{\textwidth}
  \centering
  \footnotesize
  \textsuperscript{1}Shaanxi Key Laboratory of Information Communication Network and Security, Xi'an University of Posts and Telecommunications, Xi'an, China\\
  \textsuperscript{2}School of Information Science and Technology, Northwest University, Xi'an, China\\
  \textsuperscript{3}School of Data Science and Artificial Intelligence, Chang'an University, Xi'an, China\\[0.5em]
  \textsuperscript{4}School of Computer Science and Technology, Xidian University, Xi'an, China\\
  \textsuperscript{*}Corresponding author\\
  [0.5em]
  \textbf{Correspondence:} \\
  \{\texttt{chenlong@}, \texttt{05221170@stu.}, \texttt{Mingxin Wang@stu.}\}\texttt{xupt.edu.cn}, \\
  \{\texttt{huangxiaohua}, \texttt{liuying}\}\texttt{@stumail.nwu.edu.cn}, \\
  \texttt{jiezhao@chd.edu.cn},\\
  \texttt{zyguan@xidian.edu.cn}
  \end{minipage}
}

\date{}

\begin{document}
\maketitle

\begin{abstract}
Accurate modeling of inter-stock relationships is critical for stock price forecasting. However, existing methods predominantly focus on single-state relationships, neglecting the essential complementarity between dynamic and static inter-stock relations. To solve this problem, we propose a Dual Relation Fusion Network (DRFN) to capture the long-term relative stability of stock relation structures while retaining the flexibility to respond to sudden market shifts. Our approach features a novel relative-static relation component that models time-varying long-term patterns and incorporates overnight informational influences. We capture dynamic inter-stock relationships through distance-aware mechanisms, while evolving long-term structures via recurrent fusion of dynamic relations from the prior day with the pre-defined static relations. Experiments demonstrate that our method significantly outperforms the baselines across different markets, with high sensitivity to the co-movement of relational strength and stock price. Code: \textit{https://github.com/bhx-E/DRFN}.
\end{abstract}


\section{Introduction}
Reliable stock price predictions facilitate investors in making profitable investment decisions and aid financial regulators in identifying potential financial risks. Early research on stock prediction focused on enriching the diversity of input information by introducing more market indicators \cite{mehtab2021stock} or combining texts and indicators \cite{li2020multimodal,gao2022integrating}. The subsequent studies focus on learning the temporal patterns of time-series historical data \cite{gulmez2023stock,li2024master}. However, these studies often overlook inter-stock relationships, which are critical drivers of stock price fluctuations. Accurately modeling these relationships in real-world markets remains challenging due to their inherent complexity and diversity. Specifically, they span: industry-sector correlations, vertical supply chain linkages, competitive dynamics, and cross-asset contagion effects. Existing studies learning inter-stock correlations typically adopt single-regime models, focusing exclusively on short-term fluctuations \cite{zhu2024lsr} or long-term dependencies \cite{chen2023predicting}. This leads to dual limitations: purely static models fail to capture evolving relations, while solely dynamic approaches may undermine model stability. During significant events (e.g., global crises or policy shifts), previously trained models often fail abruptly as relationships shift beyond their adaptive capacity. To this end, dynamic modeling frameworks incorporate static structural anchors to mitigate volatility. He et al. \cite{he2022static} proposed a static-dynamic graph neural network for stock recommendations. They used a data-driven graph learner to capture static/dynamic stock relations and a graph interactor to fuse multi-graph information, followed by graph convolution and ranking modules for stock recommendations. However, this study omits three critical issues: (1) News impact neglect: The effect of news on both short- and long-term stock relationships is unaddressed; (2) Time-varying structure: Stock relations evolve dynamically with new information (e.g., Apple-OpenAI's competitive-to-collaborative structural shift through strategic AI integration); and (3) Nonlinear interaction: Dynamic and static relationships exhibit complex, uncertain couplings, and simple concatenation cannot capture market complexity.

To tackle these critical shortcomings, we propose a Dual-Relation Fusion Network for stock price prediction. The key idea is to elucidate the complementarity between static and dynamic relationships, thereby enabling the model to maintain stable relationship patterns acquired over time and to rapidly adapt to abrupt changes. Specifically, to address market inefficiencies, behavioral biases, and lagged event responses — all phenomena requiring complementary multi-source information — we align news and market indicator embeddings via contrastive learning, followed by bilinear fusion. This captures the nuanced interactions between qualitative sentiment and quantitative signals, yielding robust stock representations with an enhanced ability to anticipate price movements driven by market inefficiencies. Next, we design a novel relation fusion module to capture static-dynamic complementarity through three core components: 1) Dynamic Relation Learning: Model synergistic (positive) and adversarial (negative) correlations via cross-attention with distance modulation, capturing short-term mutual influences between stocks. 2) Relative Static Relation modeling: Enhance predefined static relationships by fusing prior-day dynamic features with weighted static features, capturing time-varying long-term structures and overnight information impacts. 3) Dynamic-Relative Static Fusion: Fuse dynamic and relative-static features using bidirectional cross-attention. After relation modeling, we integrate the learned relations with the original news and market embeddings via residual connections to predict the next-day stock returns.

The main contributions are as follows: 1) Dual Relation Fusion framework: Integrates short-term volatility dynamics with relative long-term dependencies, simultaneously capturing structural stability while maintaining market responsiveness. 2) Distance-Modulated Correlation Modeling: Quantifies synergistic-adversarial stock dynamics through distance-aware mechanisms to capture time-varying inter-stock patterns. 3) Adaptive Static Relation Construction: Models evolve long-term structures via recurrent fusion of prior-day dynamic relations with the pre-defined static relations.

\section{Related Work}
Earlier studies have used historical data and technical indicators of individual stocks for predictions \cite{patel2015predicting,nelson2017stock}. Li et al. \cite{li2020multimodal} used online news text and stock price time-series market indicators as input, and they proposed a multimodal event-driven LSTM model to perform the interaction and integration of the multimodal inputs to enhance the performance of stock price prediction. However, these studies ignored correlations between stocks and deemed them independent entities, making it difficult to capture their interactive influence on stock predictions. Subsequent studies have recognized this issue and developed stock prediction methods based on relational modeling. Graph models are widely used in relationship modeling. Ying et al. \cite{ying2020time} designed a time-aware relational attention mechanism to capture the time-varying correlation strength between stocks. However, the use of predefined relations lacks an effective mechanism for handling attribute-related influences. The method in \cite{cheng2021modeling} infers latent firm relations from market signals and, with its attribute-driven graph attention network, captures attribute-sensitive momentum spillovers and complex market signal interactions. Huynh et al. \cite{huynh2023efficient} attempted to capture more sophisticated dynamic relations using industry hypergraphs, wavelet hypergraph convolutions, and temporal generative filters to capture multi-order and internal dynamic relationships. Relation modeling is further improved by capturing momentary and cross-time stock correlations through intra- and inter-stock information aggregation, using a gating mechanism for automatic feature selection based on market information \cite{li2024master}. However, this may not fully capture the long-term invariant relationships between stocks. LSR-IGRU \cite{zhu2024lsr} constructed long-short-term relationship matrices using secondary industry and overnight price information, enhancing GRU input to better integrate temporal and relationship information. Extending LSR-IGRU, we propose DRFN to capture persistent relational structures while maintaining flexibility in responding to sudden market shifts.

\begin{figure*}[t]
    \centering
        \includegraphics[scale=0.11]{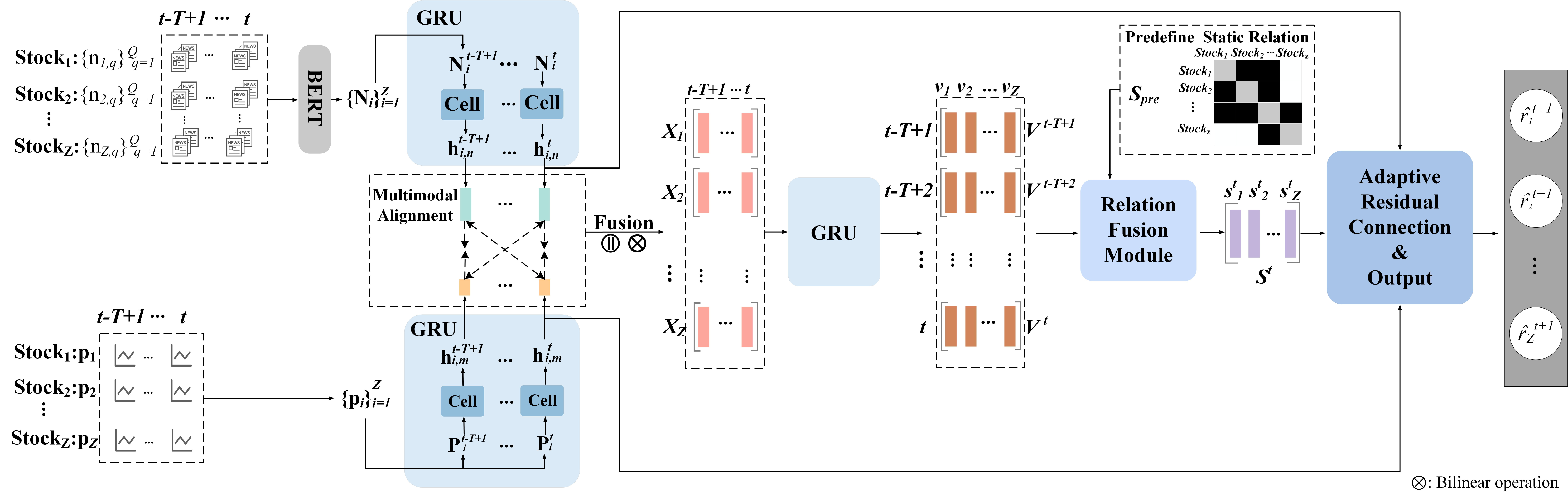}
    \caption{Overview of our framework.}
    \label{fig:System Framework Overview}
\end{figure*}

\section{Method}
We predict the next-day returns for a set of \(Z\) input stocks, formulated as a regression problem. The daily return is computed as the ratio of the interday closing price differential to the prior trading day's closing price: \(r_{t+1} = \frac{c_{t+1} - c_{t}}{c_{t}}\). For stock $i$, the input data $D = \{D_{i}^{\theta}\}^t_{\theta=t-T+1}$ contains news texts and market indicators from the past $T$ days; Specifically, on day $t$, $D_{i}^{t}$ comprises $Q$ news texts $\{ n_{i,q}^t \}_{q=1}^Q$ and six market indicators: open ($p^t_o$), close ($p^t_c$), adjusted close ($p^t_{ac}$), high ($p^t_h$), low ($p^t_l$), and volume ($p^t_v$), which form the vector $\mathbf{p}^{t}_{i} = [ p^t_{o}, p^t_{c}, p^t_{ac}, p^t_{h}, p^t_{l}, p^t_{v}]$.

The DRFN framework is illustrated in {Figure~\ref{fig:System Framework Overview}. First, we encode the input textual news and market indicator data using BERT\&GRU and GRU, respectively. The obtained temporal hidden states from both modalities are then aligned and fused to generate multimodal fusion vectors. Next, the fusion vectors are fed into the GRU module for extracting temporal patterns. Subsequently, the relation fusion module models long-term static and short-term dynamic inter-stock relations using the temporal multimodal features and the predefined static relations. Finally, the relation features are combined with the residual-propagated temporal hidden states derived from both textual news and market indicators to form unified representations. These representations are then transformed by a nonlinear mapping function for stock price predictions.

\subsection{News Text Encoding}
News sentiment drives investor behavior, triggering stock price changes. More importantly, news shapes stock correlations. For instance, supply chain news affects upstream and downstream companies; major positive news for one firm can negatively impact its competitors. We encode each news text using a pre-trained BERT model as follows:
\begin{equation}
    \mathbf{n}_{i,q}^{\theta} = BERT(n_{i,q}^{\theta}) \in \mathbb{R}^{L}
\end{equation}
The $Q$ encoded textual news embeddings in one day are represented by the matrix $\mathbf{N}_i^\theta = \left[ \mathbf{n}_{i,1}^\theta, \mathbf{n}_{i,2}^\theta, \dots, \mathbf{n}_{i,Q}^\theta \right]$. We then use a GRU to learn temporal patterns from all news over the $T$-day period:
\begin{equation}
    \mathbf{h}_{i,n}^{\theta} = GRU(\mathbf{N}_i^{\theta}, \mathbf{h}_{i,n}^{{\theta}-1}) \in \mathbb{R}^{L}
\end{equation}
where $\mathbf{h}_{i,n}^{\theta}$ represents the temporal hidden state of news embeddings at time step $\theta \in [t-T+1, t]$.

\subsection{Market Indicator Encoding}
Market indicators capture market dynamics, including trends and volatility. We model the market indicator dynamics using a separate GRU:
\begin{equation}
    \mathbf{h}_{i,m}^{\theta} = GRU(\mathbf{P}_i^{\theta}, \mathbf{h}_{i,m}^{\theta-1}) \in \mathbb{R}^{L'}
\end{equation}
where $\mathbf{h}_{i,m}^{\theta}$ denotes the temporal hidden states for the market indicators over the $T$-day time window.

\subsection{Multimodal Alignment and Fusion}
Market indicators and textual news exhibit strong complementarity, primarily grounded in Behavioral Finance. Investor decisions are systematically influenced by sentiment propagated through news narratives. Consequently, while quantitative market indicators reflect collective trading actions and outcomes, textual news reveals the qualitative narratives, sentiments, and behavioral drivers underpinning these actions. This provides a critical qualitative context that is absent from the quantitative data alone. For $Z$ stocks, we apply contrastive learning with InfoNCE loss \cite{oord2018representation} to align the temporal hidden states of the market indicators and textual news:
\begin{align}
    \begin{split}
        \mathcal{L}_{\text{Align}} = {} & -\frac{1}{T} \cdot \frac{1}{Z}\sum_{\theta=t-T+1}^{t}\sum_{i=1}^{Z} \\
        & \log \frac{\exp(\text{sim}(\mathbf{h}_{i,n}^{\theta}, \mathbf{h}_{i,m}^{\theta}) / \tau)}{\sum_{j=1}^{Z} \exp(\text{sim}(\mathbf{h}_{i,n}^{\theta}, \mathbf{h}_{j,m}^{\theta}) / \tau)}
    \end{split}
    \label{eq:modal_alignment}
\end{align}
where \( sim(\cdot) \) is the cosine similarity function; \( \tau \) denotes the temperature hyperparameter. Cross-modal alignment reduces the distance between these two modalities within the same stock, while increasing the distance across stocks between one stock's market indicator features and another stock's textual news features.

Following alignment, we use concatenation and weighted bilinear transformation to fuse the aligned cross-modal features and exploit their information complementarity:
\begin{equation}
    \mathbf{x}^{\theta}_i = tanh\left( \mathbf{W}_i [\mathbf{h}^{\theta}_{i,m} \Vert \mathbf{h}^{\theta}_{i,n}] + 
    \mathbf{h}^{\theta}_{i,m} \mathbf{T}_i^{[1:K]} \mathbf{h}^{\theta}_{i,n} + \mathbf{b}_i \right)
    \label{eq:fusion_tensor}
\end{equation}
where $\mathbf{T}_i^{[1:K]} \in \mathbb{R}^{L \times L' \times K}$ and $\mathbf{W}_i \in \mathbb{R}^{K \times (L + L')}$ are weight matrices, $\mathbf{b}_i$ is the bias term, and these parameters are shared across all time steps. $[\cdot \| \cdot]$ represents the concatenation operation. Note that we perform this fusion across all $T$ days. In fusion, concatenation preserves the original feature integrity, whereas the $K$-dimensional bilinear tensor product enables a structured cross-modal interaction. The $k$-th entry is computed using one tensor slice. Bilinear operations surpass linear combinations in modeling complex market phenomena. For instance, the impact of news sentiment on stock prices is often nonlinear, exerting greater influence beyond certain intensity thresholds, a relationship effectively captured by bilinear fusion.

\subsection{Temporal Pattern Learning}
After multimodal fusion, the sequence of fused vectors \(\{\mathbf{x}^{\theta}_i\}_{\theta=t-T+1}^{t}\) is fed into a GRU, which outputs all hidden states:
\begin{equation}
   \mathbf{v}^\theta_i = \text{GRU}\left( \{\mathbf{x}^{\theta}_i\}_{\theta=t-T+1}^{t} \right) \in \mathbb{R}^F
    \label{eq:sequential_tensor}
\end{equation}
where \(F\) denotes the hidden size of the GRU. This module captures cross-modal temporal collaborative patterns by dynamically adjusting memory retention through its update gate, focusing on the short-term impacts of news events while retaining long-term periodic patterns in market dynamics.
\begin{figure*}[t]
    \centering
        \includegraphics[scale=0.153]{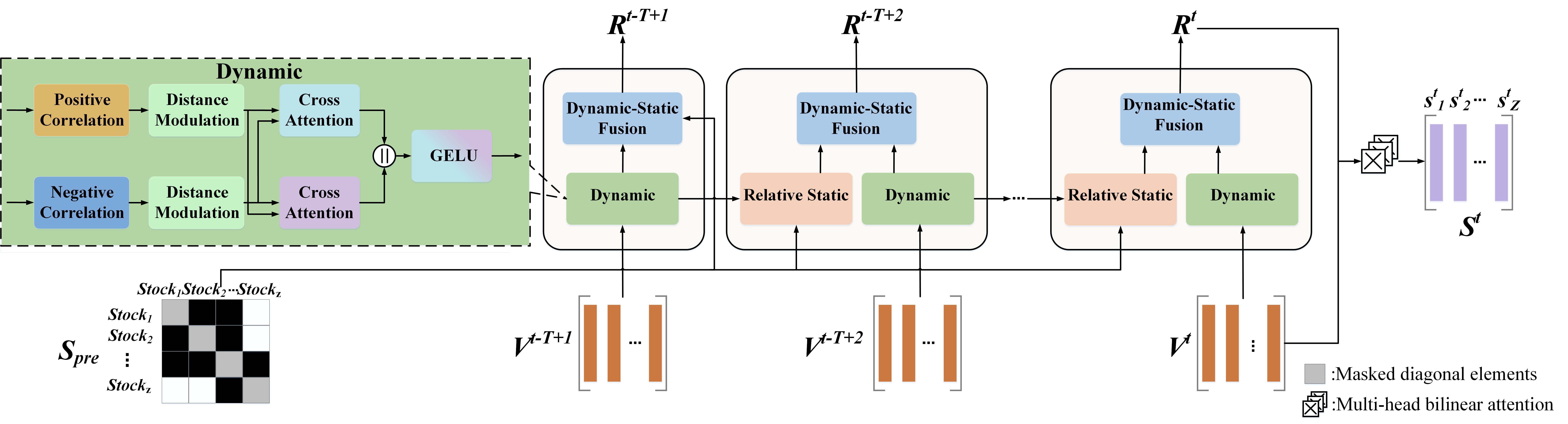}
    \caption{Relation Fusion Module}
    \label{fig:Dynamic and Relative Static Fusion}
\end{figure*}

\subsection{Relation Fusion Module}
Inter-stock relationships are not fixed and are subject to dynamic changes due to macroeconomic factors, industry, market sentiment, and other factors. To capture this complementarity, we propose a novel relation fusion module that combines short-term dynamic relationships with long-term relatively static relationships. The module structure is shown in Figure~\ref{fig:Dynamic and Relative Static Fusion}. It comprises five components: Predefined Static Relations, Dynamic Relation Learning (the zoomed-in callout in Figure~\ref{fig:Dynamic and Relative Static Fusion}), Relative-Static Relation Modeling, Dynamic-Relative Static Fusion, and Relational-Temporal Interactive Fusion.

\textbf{Predefined Static Relations.} Predefined static relationships are long-term, stable business relationships that exist among firms, derived from known market structures and enterprise connections. The predefined static relation matrix $\mathbf{S}_{pre}$ is constructed from company entities and relationships provided by Wikidata \cite{vrandevcic2014wikidata}. It is a binary matrix where 1 denotes related stocks and 0 denotes unrelated stocks, serving as the static relation structure over the past $T$-day time window. We apply a diagonal mask to $\mathbf{S}_{pre}$ to exclude self-correlation.
 
\textbf{Dynamic Relation Learning.} Stock correlations evolve dynamically under external influences. For instance, the new energy and traditional energy sectors may shift from a positive correlation during policy stimuli to a negative correlation during capital reconfiguration. To capture these time-varying inter-stock dynamics, we integrate dual-correlation modeling with distance modulation:
\begin{equation}
    \mathbf{A}^{{\theta},(a)} = \frac{\mathbf{V}^{{\theta}} \left(a \cdot \mathbf{V}^{\theta}\right)^\top}{\sqrt{F}} \in \mathbb{R}^{Z \times Z}
\end{equation}
\begin{equation}
    D^{{\theta},(a)}_{ij} = \left\| \mathbf{v}^{{\theta}}_i - a \cdot \mathbf{v}^{{\theta}}_j \right\|_1 \in \mathbb{R}^{Z \times Z}
\end{equation}
\begin{equation}
    \tilde{\mathbf{A}}^{{\theta},(a)} = \mathbf{A}^{{\theta},(a)} \odot \exp\left(-\mathbf{D}^{{\theta},(a)}\right)
\end{equation}
where $\mathbf{V}^\theta = [\mathbf{v}^\theta_1, \mathbf{v}^\theta_2, \cdots, \mathbf{v}^\theta_Z] \in \mathbb{R}^{Z \times F}$ denotes the hidden state matrix from GRU. The attention matrix $\mathbf{A}^{{\theta},(a)}$ models the relational structures among stocks at every time, with the scalar $a \in \{+1, -1\}$ switching between positive and negative correlation modes. The term $\sqrt{F}$ serves as a scaling factor to normalize the attention scores. $\mathbf{D}^{{\theta},(a)}$ quantifies pairwise behavioral discrepancies, while $\tilde{\mathbf{A}}^{{\theta},(a)}$ fuses directional alignment and distance-based modulation via Hadamard product $\odot$.
\begin{equation}
    \mathbf{F}^{{\theta},(a)} = softmax \left( \frac{ \tilde{\mathbf{A}}^{{\theta},(a)} \cdot (\tilde{\mathbf{A}}^{{\theta},(-a)})^\top }{ \sqrt{Z} } \right) \cdot \tilde{\mathbf{A}}^{{\theta},(-a)}
    \label{eq:dualatt}
\end{equation}
\begin{equation}
    \mathbf{D}_{dynamic}^{\theta} = GELU \left( \left[ \mathbf{F}^{{\theta},(+1)} \, \| \, \mathbf{F}^{\theta,(-1)} \right] \cdot \mathbf{W}_d \right)
    \label{eq:gelu}
\end{equation}
Next, cross-attention in Eq.~\ref{eq:dualatt} performs cross-perspective fusion by: 1) Identifying stock pairs exhibiting strong positive but weak negative correlations (or vice versa) through interactions between $\tilde{\mathbf{A}}^{\theta,(a)}$ and $\tilde{\mathbf{A}}^{\theta,(-a)}$;  2) Assigning cross-perspective importance weights via softmax to highlight dominant association patterns. Then a nonlinear projection with a learnable parameter matrix $\mathbf{W}_d \in \mathbb{R}^{2Z \times Z}$ integrates the positive and negative correlation fusion matrices $\mathbf{F}^{\theta,(+1)}$ and $\mathbf{F}^{\theta,(-1)}$ to produce the dynamic relation matrix $\mathbf{D}_{\text{dynamic}}^{\theta} \in \mathbb{R}^{Z \times Z}$, which dynamically evolves with market conditions.

\textbf{Relative-Static Relation Modeling.} Overnight information is released intensively after the opening of the next trading day and has a significant impact on stock price movements \cite{lou2019tug}. Nevertheless, predefined static relations solely capture long-term relation patterns while neglecting this ultra-short-term information. To address this limitation, we integrate the predefined static relations with the dynamic relation from the preceding day, resulting in a final static relation termed the relative-static relation. Specifically, we employ linear weighting to combine the dynamic relation of the previous time step \(\mathbf{D}_{dynamic}^{\theta-1} \in \mathbb{R}^{Z \times Z}\) and the predefined static relation \(\mathbf{S}_{pre} \in \mathbb{R}^{Z \times Z}\). The computational formula is as follows:
\begin{equation}
    \mathbf{S}_{relative}^{\theta} = \alpha\cdot\mathbf{D}_{dynamic}^{\theta-1}+(1-\alpha) \cdot [\mathbf{W}_s\odot\mathbf{S}_{pre}]
    \label{eq:Relative Static}
\end{equation}
where \(\mathbf{S}_{relative}^{\theta} \in \mathbb{R}^{Z \times Z}\) is the output at time step $\theta$. We add a learnable parameter matrix $\mathbf{W}_s$ that is reused across all time steps to dynamically adjust \(\mathbf{S}_{pre}\). This adjustment enables the predefined static relation to capture the time-varying characteristics of the relationship strength among stocks while maintaining a stable long-term relation structure. Besides, we introduce a learnable scalar parameter $\alpha$ that determines the relative contributions of dynamic and static relationships. Note that at the initial time step $t-T+1$, \(\mathbf{S}_{pre}\) serves as the static relation component since prior dynamic relations are unavailable.

\textbf{Dynamic-Relative Static Fusion.} To capture the complementarity between dynamic and static relationships, we design a dual-relation fusion mechanism that generates comprehensive representations reflecting both short-term fluctuations and long-term stability. Real market conditions reveal no fixed proportionality between these relationships; instead, they present complex and uncertain interactions. Therefore, we implement bidirectional cross-attention for adaptive fusion, enabling dynamic-static integration through bidirectional information flow:

\begin{align}
\mathbf{R}^{\theta} &= GELU\Big( \Big[ \text{Attention}\left( \mathbf{S}^{\theta}_{\text{relative}} ,\mathbf{D}^{\theta}_{\text{dynamic}}, \mathbf{D}^{\theta}_{\text{dynamic}} \right) \notag \\
&\quad \parallel\ \text{Attention}\left( \mathbf{D}^{\theta}_{\text{dynamic}}, \mathbf{S}^{\theta}_{\text{relative}}, \mathbf{S}^{\theta}_{\text{relative}} \right) \Big] \cdot \mathbf{W}_{DRS}  \Big) 
\label{eq:Dynamic and Static}
\end{align}
\begin{equation}
\text{Attention}(Q, K, V) = Softmax\left( \frac{QK^\top}{\sqrt{d_k}} \right) V
\end{equation}
where $\mathbf{R}^{\theta}$ is the fused representation at step $\theta$, and $\mathbf{W}_{DRS} \in \mathbb{R}^{2Z \times Z}$ is a learnable weight matrix. Unlike traditional fusion methods, our bidirectional cross-attention mechanism adaptively controls the degree of fusion between relationships through its inherent adaptability, thereby effectively exploiting their complementary synergy.

\textbf{Relational-Temporal Interactive Fusion.} Building on the stock relations, we further fuse their temporal multimodal features by applying a multi-head bilinear attention mechanism to the relation matrix \(\mathbf{R}^{t}\) and the temporal multimodal feature matrix \(\mathbf{V}^t\):
\begin{equation}
\mathbf{S}^t = \underset{b=1}{\overset{B}{\|}}  tanh\left( \left[ \mathbf{R}^{t,[b]} \cdot (\mathbf{V}^t \mathbf{W}_{TR}^{[b]}) \right] \right) \in R^{Z \times BF'}
\label{eq:Comprehensive Relation}
\end{equation}
where \(\mathbf{S}^t = [\mathbf{s}^t_1, \mathbf{s}^t_2, \cdots, \mathbf{s}^t_Z]\) denotes the final fused representation at time step \(t\), the terminal point of the \(T\)-day time window. $\mathbf{W}_{TR}^{[b]} \in R^{F \times F'}$ denotes the interaction weight matrix of the $b$-th attention head. This attention mechanism guides the model in learning the interaction between relationships and time by exploring multiple feature subspaces. For instance, a stock may be competitive or complementary to other stocks, and these relationships may be reflected in time series as synergies or differences in characteristics such as stock price volatility and trading volume changes.

\subsection{Adaptive Residual Connection and Output}
To capture the immediate impact of news and indicators, we employ residual connections and adaptively fuse temporal embeddings with relation features to generate a unified representation. The fused representation is subsequently transformed via a nonlinear projection to predict stock returns. The final output is computed as:
\begin{align}
\hat{r}^{t+1}_i = \tanh\Big( \mathbf{W_o} \cdot [&\, (\mathbf{s}^t_i)^\top 
+ \alpha_i^{(1)} \cdot (\mathbf{W}_n \cdot(\mathbf{h}_{i,n}^t)^\top + \mathbf{b}_n) \notag \\
&+ \alpha_i^{(2)} \cdot (\mathbf{W}_m \cdot (\mathbf{h}_{i,m}^t)^\top + \mathbf{b}_m)] + b_o \Big)
\label{eq:output_fusion}
\end{align}
where $\mathbf{W}_o \in \mathbb{R}^{1 \times BF'}$, $\mathbf{W}_n \in \mathbb{R}^{BF' \times L}$, and $\mathbf{W}_m \in \mathbb{R}^{BF' \times L'}$ denote weight matrices, with corresponding biases $b_o \in \mathbb{R}$, $\mathbf{b}_n \in \mathbb{R}^{BF'}$, and $\mathbf{b}_m \in \mathbb{R}^{BF'}$. The scalars $\alpha_i^{(1)}$ and $\alpha_i^{(2)}$ are learnable parameters. This adaptive fusion addresses two key challenges: 1) Feature Heterogeneity: Input features are diverse, making simple concatenation noisy. 2) Dynamic Sensitivity: Stocks exhibit varying sensitivities (e.g., policy-sensitive stocks rely more on news, while energy stocks focus more on crude oil prices). Learnable weights enable dynamic parameter adjustment, thus balancing the contributions of different information sources.
\subsection{Objective Function}
We combine Mean Squared Error (MSE) loss and cross-modal alignment loss as the overall objective function:
\begin{equation}
\mathcal{L} = \mathcal{L}_{\text{MSE}} + \lambda \mathcal{L}_{\text{Align}}
\label{eq:total_loss}
\end{equation}
\begin{equation}
\mathcal{L}_{\text{MSE}} = \frac{1}{Z} \sum_{i=1}^{Z} \left( r_i^{t+1} - \hat{r}_i^{t+1} \right)^2
\label{eq:mse_loss}
\end{equation}
where $r_i^{t+1}$ and $\hat{r}_i^{t+1}$ denote the ground-truth and predicted returns of stock $i$ at time $t+1$, respectively, and $\lambda$ is a balancing hyperparameter.

\section{Experiments}

\begin{table*}[t]
\centering
\begin{tabular}{c|c|c|c|
>{\centering\arraybackslash}p{2cm}|
>{\centering\arraybackslash}p{2cm}|
>{\centering\arraybackslash}p{2cm}|
>{\centering\arraybackslash}p{2cm}}

\toprule
\multirow{2}{*}{Model} & \multirow{2}{*}{Relation} & \multirow{2}{*}{Text} & \multirow{2}{*}{Price} & \multicolumn{2}{c|}{CMIN-US} & \multicolumn{2}{c}{CMIN-CN}  \\
&   &   &    & RMSE (Avg.)    & MAE (Avg.)     & RMSE (Avg.)   & MAE (Avg.)      \\
\midrule
LSTM  & \texttimes & \checkmark & \checkmark  &  0.0232   &0.0180     &0.0359    &0.0285   \\
GRU   & \texttimes  & \checkmark & \checkmark & 0.0230    &0.0177     &0.0356    & 0.0277   \\
GCN   & \checkmark & \checkmark & \checkmark & 0.0221     & 0.0168    &0.0347    & 0.0264        \\
GAT   & \checkmark & \checkmark & \checkmark & 0.0217   &0.0164   & 0.0343   &0.0257  \\
MASTER & \checkmark & \texttimes & \checkmark   &0.0205    &  0.0156   & 0.0331   & 0.0245  \\
SAMBA & \texttimes & \texttimes & \checkmark   & 0.0201   &0.0153    & 0.0325   & 0.0237     \\
LSR-IGRU & \checkmark  & \texttimes & \checkmark &0.0192     & 0.0141    &0.0317     &0.0228      \\
\textbf{DRFN} & \checkmark  &\checkmark &  \checkmark & \textbf{0.0166} & \textbf{0.0117} & \textbf{0.0270} & \textbf{0.0191} \\  
\bottomrule
\end{tabular}
\caption{Main comparison results}
\label{tab:a}
\end{table*}

\subsection{Dataset and Experiment Settings}
We use datasets from China (CMIN-CN) and the US (CMIN-US) markets~\cite{luo2023causality} for evaluation. CMIN-US contains 105 stocks over 1,006 trading days; CMIN-CN comprises 227 stocks across 968 days\footnote{Samples from missing trading days are removed.}. Both datasets are chronologically split 8:1:1 into training, validation, and test sets. We use the Root Mean Square Error (RMSE) and Mean Absolute Error (MAE) as evaluation metrics. \emph{Due to market volatility, each baseline is trained 10 times with random initialization, and the average performance is reported.}

Cross-market hyperparameter tuning produced identical configurations on both validation sets, demonstrating strong cross-market generalization. Data specifications: maximum daily news count $Q=30$ (40 tokens/text); BERT-encoded text vectors $L=768$ (GRU preserves dimension); GRU-encoded temporal vectors of market indicators $L'=64$. Optimal model parameters: time window $T=5$ ($T \in [2,10]$), attention heads $B=6$ ($B \in \{2,4,6,8,10\}$) using even counts for symmetric modeling. Dimensions: GRU $F=128$ and attention $F'=256$ (both $\in \{32,64,128,256,512\}$), balance parameter $\lambda=1$ ($\lambda \in \{0.1,0.5,1,1.5,2\}$). The model was trained using the Adam optimizer with a learning rate of \(2 \times 10^{-3}\), implemented on a single NVIDIA Tesla P100 GPU (16G), with a total runtime of approximately 11 hours for convergence.

\subsection{Baselines and Main Comparisons}
\textbf{LSTM and GRU.} Long Short-Term Memory (LSTM) \cite{hochreiter1997long} and Gated Recurrent Unit (GRU) \cite{chung2014empirical} are classic sequence models widely used for time-series prediction tasks. We employ a 2-layer LSTM and 2-layer GRU as sequence-model-based baseline methods.

\noindent\textbf{GAT.} Graph Attention neTwork \cite{velickovic2017graph}  assigns varying weights to different nodes within a neighborhood. We use GAT with two masked self-attention layers.

\noindent\textbf{GCN.} The Graph Convolutional Network \cite{kipf2017semi} is commonly used to capture stock relations. We employ a GCN with two convolutional layers as a baseline. The attributes of adjacent nodes are linearly aggregated to the central node using normalized stock relationships. 

\noindent\textbf{SAMBA.} \cite{mehrabian2025mamba} proposed SAMBA, a framework that combines the Mamba architecture and graph neural networks (GNNs). SAMBA employs a bidirectional Mamba module to capture long-term dependencies in historical price data and to learn inter-stock-feature dependencies via adaptive graph convolution.

\noindent\textbf{MASTER.} {\cite{li2024master} proposed a Market-Guided Stock TransformER model to capture momentary and cross-time correlations among stocks using market indicators.

\noindent\textbf{LSR-IGRU.} \cite{zhu2024lsr} proposed an Improved GRU to capture the Long Short-term Relations of stocks. It uses secondary industry information and overnight price changes to construct long- and short-term stock relation representations, and then the improved GRU fuses the original input with these representations via concatenating and weighting unit hidden states.

\noindent\textbf{DRFN.} This is our method.

\noindent Note that LSTM and GRU use concatenation of technical indicators and text features as inputs. GCN and GAT employ indicator and text feature combinations as graph nodes and connect them with predefined static relations.

The comparison results for the two markets are similar, shown in Table \ref{tab:a}. LSTM and GRU achieve the lowest performance because they capture only fundamental temporal patterns, which are insufficient for modeling underlying relationships in multimodal data. GAT and GCN surpass the two basic sequence models because graph-based architectures effectively model the stock relationships. In comparison, MASTER achieves a better performance by capturing fine-grained stock relations, both intra- and inter-stock, along with asymmetric and dynamic inter-stock correlations. SAMBA outperforms MASTER, due to its enhanced capture of long-term dependencies via bidirectional time-series modeling and its effective modeling of intra-stock feature dependencies through adaptive structural modeling. LSR-IGRU outperforms SAMBA through its incorporation of primary/secondary industry data and overnight prices in relation modeling. This approach effectively captures cross-industry dependencies and short-term price fluctuation patterns, yielding superior results through richer market-signal extraction. By leveraging complementary learning of static and dynamic stock relations, our method achieves optimal performance across both markets.

\begin{table*}[t]
\centering
\begin{tabular}{c|
>{\centering\arraybackslash}p{2cm}|
>{\centering\arraybackslash}p{2cm}|
>{\centering\arraybackslash}p{2cm}|
>{\centering\arraybackslash}p{2cm}} 
\toprule
\multirow{2}{*}{Model} & \multicolumn{2}{c|}{CMIN-US} & \multicolumn{2}{c}{CMIN-CN}  \\
& RMSE (Avg.)    & MAE (Avg.)      & RMSE (Avg.)   & MAE (Avg.)    \\
\midrule
Static & 0.0221   &0.0162    & 0.0337  &0.0250     \\
Relative Static & 0.0208  & 0.0151    & 0.0326   & 0.0240   \\
Dynamic  & 0.0196   &0.0145    &0.0315    &0.0233     \\
w/o Modality Alignment  & 0.0191   &0.0139     &0.0310    &0.0225    \\
w/o adaptive weighted residual mechanism  & 0.0174   & 0.0123    & 0.0280   &0.0204    \\
\textbf{DRFN} & \textbf{0.0166} & \textbf{0.0117} & \textbf{0.0270} & \textbf{0.0191} \\  
\bottomrule
\end{tabular}
\caption{Ablation study}
\label{tab:b}
\end{table*}

\subsection{Relation Modeling Variants and Ablation Study}
Considering the centrality of relation modeling to our framework, we evaluate the relation fusion module against its variants to demonstrate its effectiveness, with complementary ablation studies of other components. The results are shown in Table \ref{tab:b}. Replacing the relation fusion module with Predefined Static Relations (\emph{Static}) causes a significant performance deterioration. This demonstrates that stock relations are inherently dynamic. \emph{Relative Static} employs the Relative Static Relating Modeling, outperforming \emph{Static} by capturing time-varying patterns of the stocks' relations in long-term trading ranges and incorporating overnight information. \emph{Dynamic} (using only Dynamic Relation Learning) performs best among the three variants, proving that short-term dynamic stock relations correlate more strongly with price fluctuations than long-term static relations. Nevertheless, it consistently underperforms our static-dynamic fusion method because static structural anchors are essential in dynamic modeling frameworks to mitigate market volatility. Further ablation studies highlight modality alignment's necessity and the direct influence of input embeddings on stock dynamics.

\subsection{Relation Attention Sensitivity to Price Fluctuations}

\begin{figure}[htbp]
    \centering
    \includegraphics[width=0.6\textwidth]{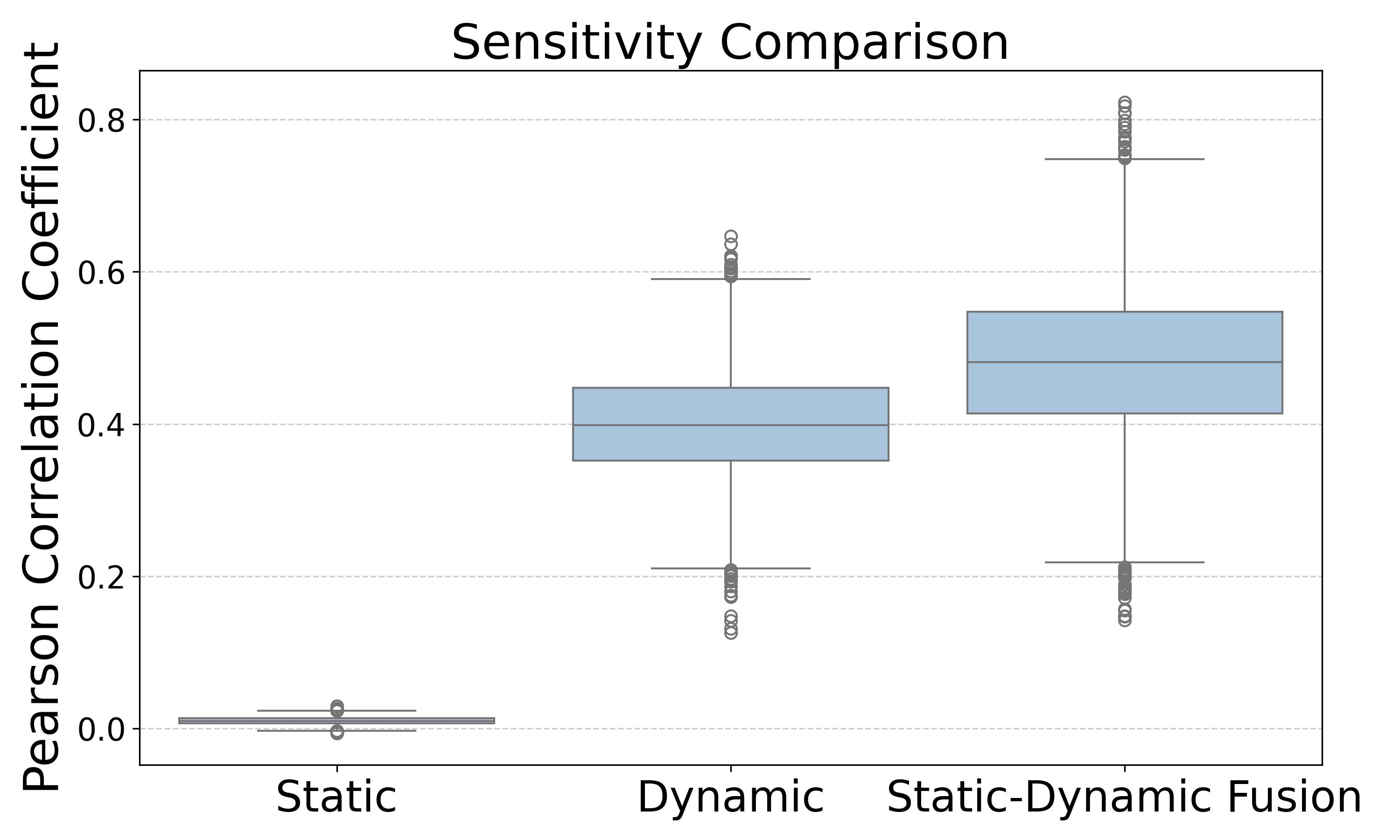}  
    \caption{Sensitivity analysis}
    \label{fig:sensitivity_analysis}
\end{figure}

\begin{figure}[htbp]
    \centering
    \includegraphics[width=0.9\textwidth]{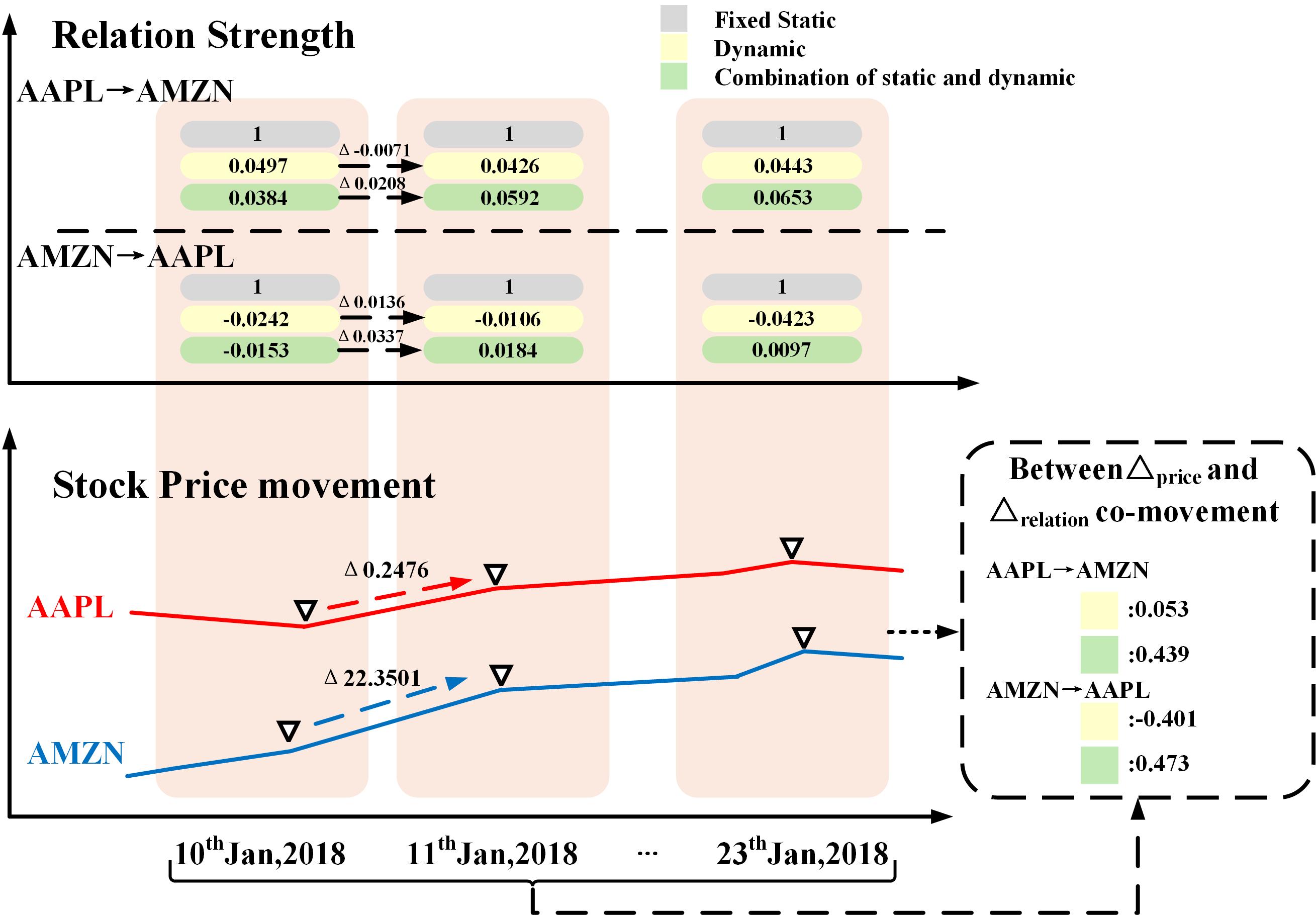}  
    \caption{Comparative sensitivity of different relation modeling methods to AAPL and AMZN price movements}
    \label{fig:AAPL_AMZN_analysis}
\end{figure}
Modeling stock relationships is the main contribution of our work. We evaluate whether attention intensity between stocks in the relational module responds to price fluctuations. Specifically, we assess the sensitivity of three relational modeling approaches to market turbulence on the test set\footnote{Given consistent cross-market comparison results, we analyze the full 90-trading-day test set of US market data in CMIN-US.}: 1) the Static Relation method employed in GAT \cite{velickovic2017graph}, 2) the Dynamic Relation Learning proposed in LSR-IGRU \cite{zhu2024lsr}, and 3) our Static-Dynamic Fusion. To measure stock relationship sensitivity to price fluctuations, we compute the Pearson correlation between relational strength changes ($\Delta S$) and price changes ($\Delta C$):
\begin{equation}
\rho = \frac{ \sum_{\theta=1}^{T'} \left( \Delta S^{(\theta)} - \overline{\Delta S} \right) \left( \Delta C^{(\theta)} - \overline{\Delta C} \right) }
{ \sqrt{ \sum_{\theta=1}^{T'} \left( \Delta S^{(\theta)} - \overline{\Delta S} \right)^2 } \cdot \sqrt{ \sum_{\theta=1}^{T'} \left( \Delta C^{(\theta)} - \overline{\Delta C} \right)^2 } }
\end{equation}
where $\Delta S^{(\theta)}$ and $\Delta C^{(\theta)}$ are the predicted changes in relational strength and stock price from $\theta-1$ to $\theta$, with $\overline{\Delta S}$ and $\overline{\Delta C}$ their respective means over 90-day windows ($T' = 90$). The results are shown in {Figure~\ref{fig:sensitivity_comparison Analysis}}. Static relation modeling with predefined stock relations inherently lacks price sensitivity, owing to its fixed connectivity. While modeling dynamic stock relations adapts inter-stock connectivity to market fluctuations, its relational sensitivity to price volatility remains minimal. This is why LSR-IGRU underperforms our method because it models stock relationships through overnight price similarity and historical patterns but fails to capture real-time relationship dynamics driven by price fluctuations, a fundamental limitation that causes performance deterioration. Our Static-Dynamic Fusion module achieves superior sensitivity to relation-price co-movement. By recurrently integrating historical and current dynamics, we accumulate cross-period relations, thereby enhancing co-movement sensitivity.

We further compare the sensitivity of different relation modeling methods to market shifts by visualizing the relational strength and stock prices for AAPL and AMZN (January 10--23, 2018) in Figure~\ref{fig:AAPL and AMZN Example Analysis}. Long-term analysis reveals a weak correlation between the dynamic module and actual price fluctuations (coefficients: 0.053 and -0.401), particularly in directional consistency. In stark contrast, our Static-Dynamic Fusion module achieves significantly stronger correlations (0.439 and 0.473). A Short-term analysis of January 10-11 is a representative case: During synchronous AAPL-AMZN price surges, our Static-Dynamic Fusion module closely tracked price movements in both direction and amplitude, confirming strong short-term synergy; the dynamic module, however, failed to capture this co-movement. Building on our quantitative analysis, the case study visually illustrates enhanced relation-to-price sensitivity in our static-dynamic fusion module, achieved through temporal superposition effects of dynamic relations.

\section{Conclusions}
In this study, we propose a dual-relation fusion network to achieve complementary fusion between static and dynamic stock relations, resulting in the learning of the long-term relative stability of the stock relation structure while maintaining the model's flexibility in responding to sudden market fluctuations. Our method demonstrates consistent superiority in comparative experiments, with the relation fusion module exhibiting enhanced sensitivity to temporal stock price fluctuations.

\section{Acknowledgments}
This paper is supported by the National Natural Science Foundation of China (62203354, 62472340, 6230336, 62133012, 62425605, 62306235, 62502051), the Key Research and Development Program of Shaanxi (Program Nos. 2024GX-YBXM-122, 22025CY-YBXM-041, 2024CY2-GJHX-15, 2022ZDLGY01-10), the Natural Science Basic Research Program of Shaanxi (Program Nos. 2025JC-QYXQ-040, 2025JC-YBQN-900, 2025CY-YBXM-067), Guangdong Basic and Applied Basic Research Foundation (Program No.2025A1515010267), the Postdoctoral Research Project Sustentation Fund of Shaanxi (No.2024BSHSDZZ112), the Postdoctoral Fellowship Program of CPSF (Grant No. GZC20251101) and the Key Scientific Research Projects of the Shaanxi Provincial Education Department (No. 24JR152).

\bibliographystyle{unsrt}  

\bibliography{main} 

\end{document}